%% file: root.tex
\documentclass[letterpaper, 10 pt, conference]{ieeeconf}  

\IEEEoverridecommandlockouts                              

\usepackage{graphics} 
\usepackage{amsmath} 
\usepackage{amssymb}  
\usepackage{bm}
\usepackage{algorithmic}
\usepackage{cleveref}[capitalize]
\usepackage{caption}
\usepackage{subcaption}
\usepackage{siunitx}
\usepackage{tikz}
\usepackage{pgfplots}
\usepgfplotslibrary{fillbetween}
\usetikzlibrary{positioning, calc, fit}
\usetikzlibrary{patterns, shapes.geometric, arrows.meta, fit, 3d, angles,quotes}
\usetikzlibrary{shapes.misc}
\usepackage{url}

\title{\LARGE \bf
Sample-Efficient Learning to Solve a Real-World Labyrinth Game Using Data-Augmented Model-Based Reinforcement Learning
}

\author{Thomas Bi$^{1}$ and Raffaello D'Andrea$^{1}$
	\thanks{$^{1}$ Institute for Dynamic Systems and Control, ETH Zurich, Switerland, {\tt\small bit@ethz.ch}}%
}

\begin{document}

\maketitle
\thispagestyle{empty}
\pagestyle{empty}

\begin{abstract}

Motivated by the challenge of achieving rapid learning in physical environments, this paper presents the development and training of a robotic system designed to navigate and solve a labyrinth game using model-based reinforcement learning techniques. The method involves extracting low-dimensional observations from camera images, along with a cropped and rectified image patch centered on the current position within the labyrinth, providing valuable information about the labyrinth layout. The learning of a control policy is performed purely on the physical system using model-based reinforcement learning, where the progress along the labyrinth's path serves as a reward signal. Additionally, we exploit the system's inherent symmetries to augment the training data. Consequently, our approach learns to successfully solve a popular real-world labyrinth game in record time, with only 5 hours of real-world training data.

\end{abstract}

\input{sections/introduction}
\input{sections/hardware}

\input{sections/method}
\input{sections/results}

\input{sections/conclusion}






\section*{ACKNOWLEDGMENT}

The authors would like to thank Daniel Wagner, Matthias Mueller, and Tim Flueckiger for their contributions to the development of the system, and Pete Wurman for fruitful discussions.




\bibliographystyle{IEEEtran}
\bibliography{references}

\end{document}

%% file: sections/introduction.tex
\section{INTRODUCTION}
\label{sec:intro}

\begin{figure}
	\centering
	\includegraphics[width=0.83\columnwidth]{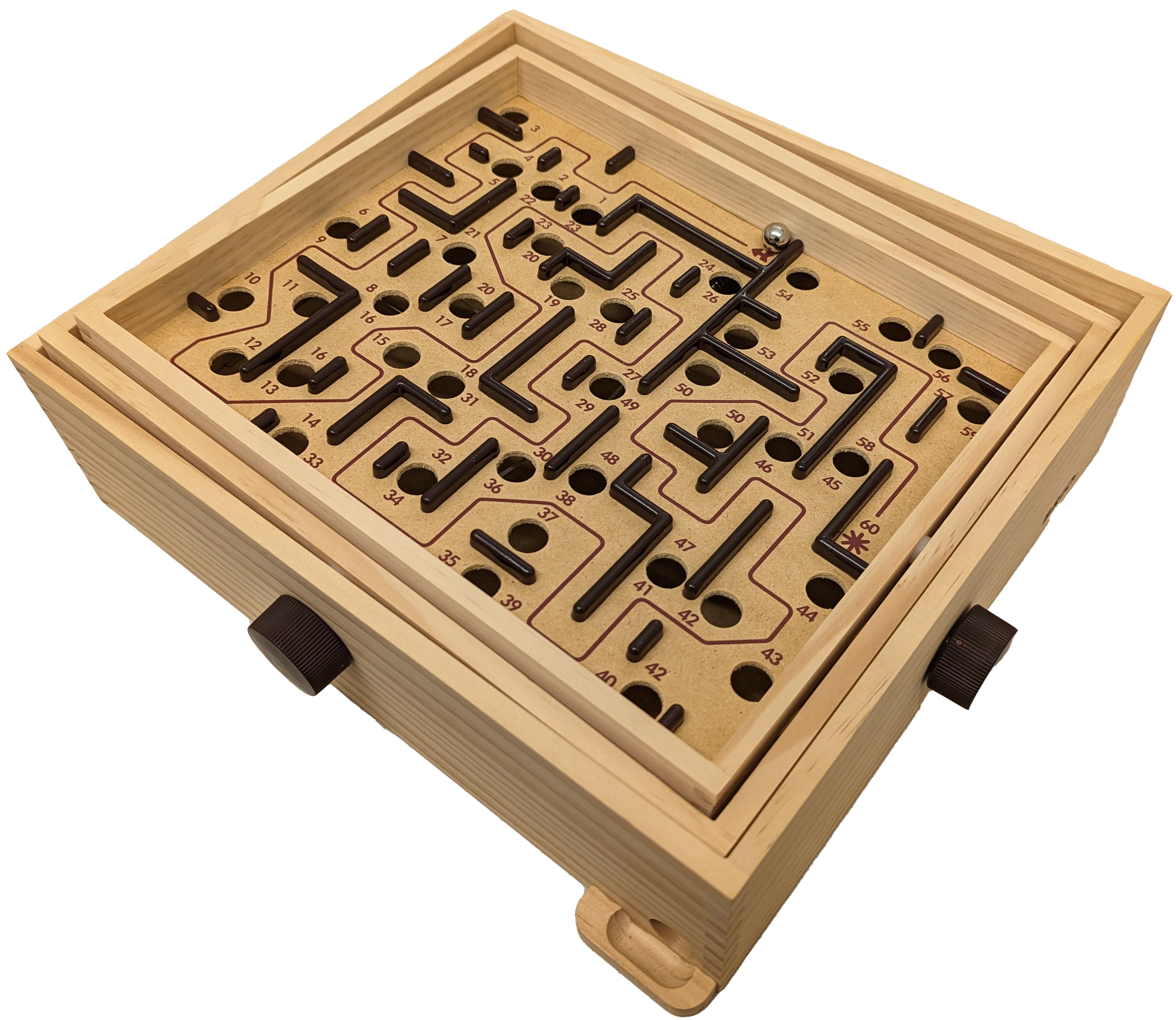}
	\caption{The labyrinth game is a marble game with the goal of steering a ball from a start to an end position while avoiding letting the ball fall down the holes. Pictured above is the BRIO Labyrinth, introduced almost 80 years ago, and with millions sold.}
	\label{fig:brio}
\end{figure}

The labyrinth game, pictured in \Cref{fig:brio}, is a marble game consisting of a maze with walls and holes. The objective is to steer a steel ball from a given start point to the end point by tilting the playfield using two knobs. In doing so, the player must prevent the ball from falling down any of the holes.
While it is a relatively straightforward game, it requires fine motor skills and spatial reasoning abilities, and, from experience, humans require a great amount of practice to become proficient at the game. Moreover, the labyrinth presents additional challenges, including stiction effects between the ball and the walls/floor, irregularities in the labyrinth's surface, and the presence of nonlinear coupling between the control knobs and the labyrinth board. Due to these reasons, we believe that the labyrinth provides an excellent platform to apply and evaluate state-of-the-art robotic learning approaches, and advance our understanding of fine motor control, spatial reasoning, and learning for dynamic systems.

In this paper, we present a method that efficiently learns to solve a popular labyrinth game, the BRIO labyrinth, faster than any previously recorded time (to the best of our knowledge), using only $5$ hours of experience collected on the physical system. The following aspects of our method are key to its success. First, observations that efficiently encode the relevant information necessary to infer the optimal strategy are extracted from images captured by a camera positioned above the labyrinth. These observations include the ball position, plate inclination angles, directional information about the path through the labyrinth, and a rectified image patch centered around the current ball position. This image patch provides information about the presence of walls and holes in the vicinity of the ball which can be used by the learned policy to efficiently avoid holes and exploit the collision of the ball with walls to redirect the ball. Second, a dense reward function is defined based on the progress through the labyrinth. 
Lastly, we combine the state-of-the-art model-based reinforcement learning (RL) approach DreamerV3\cite{hafner2023dreamerv3} with a data augmentation technique, leveraging the system's presumed symmetry. Specifically, during training, we randomly mirror the observations about two planes of symmetry, thereby generating more diverse training data and improving the generalization to unseen parts of the labyrinth during training.

\subsection{Related work}

In recent years, deep RL has excelled at attaining super-human performance amongst a variety of different games such as chess, Go, and shogi \cite{schrittwieser2020mastering}, Atari video games \cite{burda2018large}, 2D and 3D navigation in mazes (\cite{juliani2020, devo2020deep, parisotto2017neural}), Minecraft \cite{baker2022video}, and Gran Turismo \cite{wurman2022outracing}. 
However, similar success in systems that require physical interactions with the real world is limited; examples include autonomous drones that have been able to beat human world champions at drone racing \cite{kaufmann2023champion}, and a curling robot that reached the level of top-ranked women's teams \cite{won2020adaptive}. The few examples that do exist require expensive and bespoke hardware and are thus very difficult to reproduce, let alone exploit.

To further study the potential of artificial intelligence in the real world, we believe the labyrinth game to be an ideal testbed due to its cost-efficiency, wide availability, and challenging gameplay. Previous research includes \cite{frid2020path}, where a model-based path-following controller is employed to solve a real-world labyrinth, whereby the inclination angles of the board are controlled by two servos, and a camera tracks the ball position. In \cite{ofjall2012combining}, the inverse dynamics of the real-world board are learned using the mixture of several locally weighted linear models, where the parameters of each model are learned online. In order to succeed, the two mentioned works both replace the original labyrinth with custom boards with simpler layouts and attempt to remove any unevenness from the surface. 
In contrast, the authors of \cite{andersen1993real} do not make any direct modifications to the game and aim to solve it in its original state. A camera captures the position of the ball and the inclination angles of the board. A Kalman Filter \cite{kalman1960new} and state-feedback controller are then designed based on a linear ball-and-plate model. The resulting closed-loop system is able to successfully complete the labyrinth with a 75\% success rate.
Approaches employing RL to solve the labyrinth include \cite{abdenebaoui2007connectionist} and \cite{metzen2009brio}. However, the validity of these approaches was only verified in simulation.

In this work, we exploit recent advances in model-based RL and its ability to make informed decisions about potentially successful behaviours by planning into the future. The effectiveness of learned models has previously been demonstrated by the authors of \cite{wu2023daydreamer}, where world models enable a physical quadrupedal robot to learn to stand up and walk within 1 hour. By leveraging such an approach we obtain a highly performant policy that solves a popular real-world labyrinth game, the BRIO Labyrinth, in record time. In a similar spirit to the work in \cite{andersen1993real}, we do not modify the game itself and use the original labyrinth without modifying any of its dynamics. 


\subsection{Outline}

The physical setup is presented in \Cref{sec:hardware}. In \Cref{sec:method}, the proposed method and its components are described. The performance of the policy learned on the real-world system, as well as an ablation study, are presented in \Cref{sec:results}. Finally, \Cref{sec:conc} draws conclusions and gives an outlook on future work. 

%% file: sections/hardware.tex
\section{Hardware}
\label{sec:hardware}

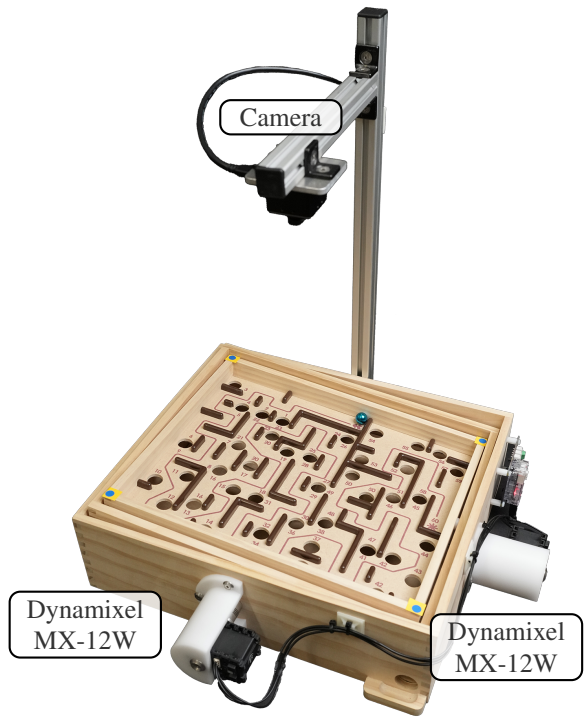
\begin{figure}
    \centering
    \begin{tikzpicture}
        \node at (0, 0) {\includegraphics[width=0.8\columnwidth]{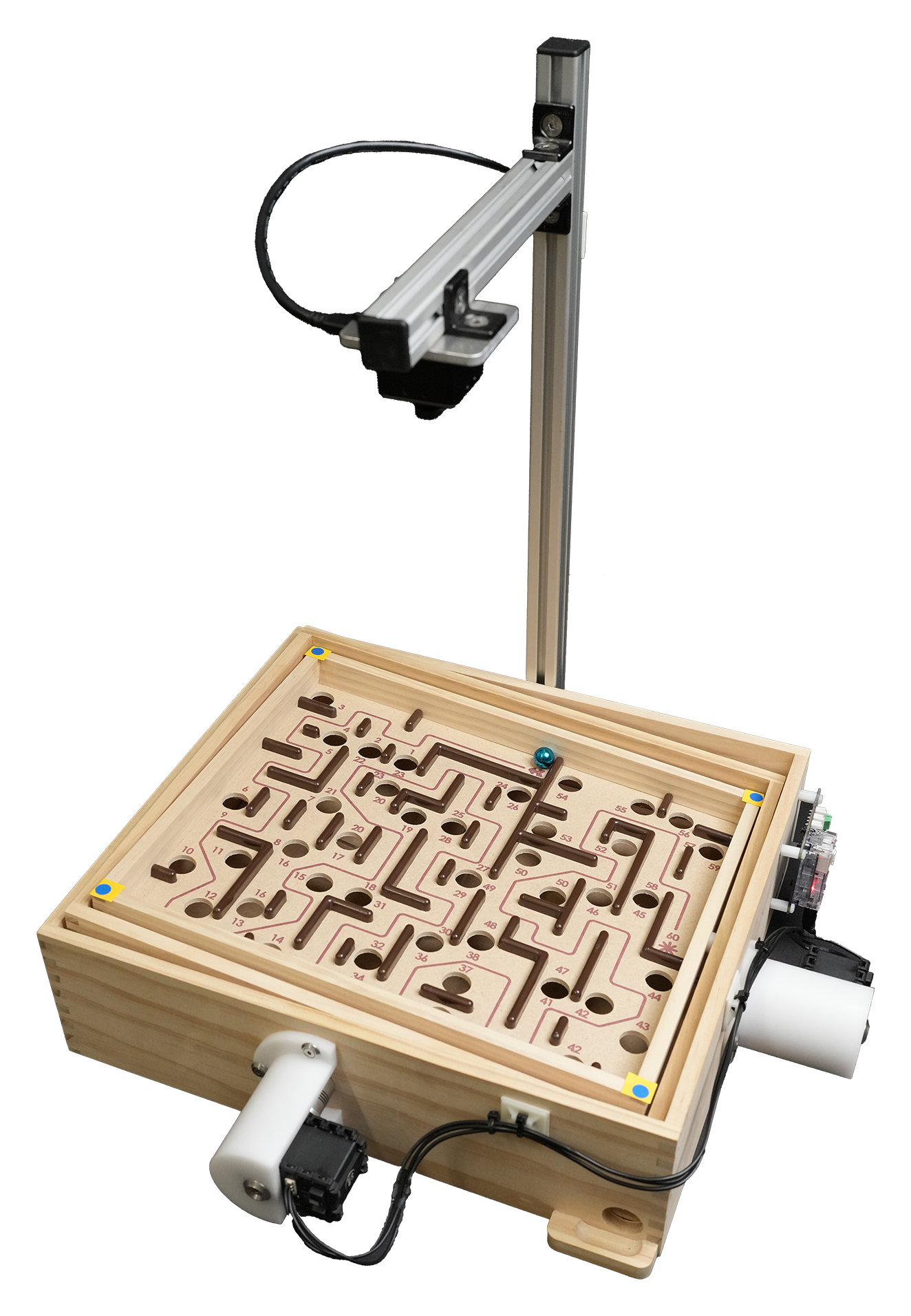}};
        \node[text width=1.8cm, inner sep=3pt, align=center, draw, thick, fill=black!0, fill opacity=0.80, rectangle, rounded corners] at (2.6, -3.8) {Dynamixel MX-12W};
        \node[text width=1.8cm, inner sep=3pt, align=center, draw, thick, fill=black!0, fill opacity=0.80, rectangle, rounded corners] at (-3.0, -3.5) {Dynamixel MX-12W};
        \node[text width=1.4cm, inner sep=3pt, align=center, draw, thick, fill=black!0, fill opacity=0.80, rectangle, rounded corners] at (-0.4, 3.25) {Camera};
    \end{tikzpicture}
    \caption{The physical setup of our system. Two Dynamixel MX-12W motors are directly coupled to the shafts controlling the inclination angles of the labyrinth plate, while a wide-angle USB3 camera captures top-down images of the labyrinth.}
    \label{fig:hardware}
\end{figure}

The robotic system considered in this paper consists of four main parts; (i) the labyrinth itself, (ii) two motors that control the inclination angles of the labyrinth plate, (iii) a wide-angle camera capturing a full view of the system, and finally, (iv) a workstation that concurrently runs all the presented algorithms. The system is pictured in \Cref{fig:hardware}.

\subsection{Motors}

In order to control the orientation of the labyrinth board, the original BRIO labyrinth has two plastic knobs press fitted onto two shafts which control the orientation around the two axes. We remove the knobs, couple two Dynamixel MX-12W motors directly to the two shafts, and mount the motors to the sides of the labyrinth.
Furthermore, the motors are daisy-chained and connected to a Dynamixel U2D2 USB converter which allows for commanding the motors using serial communication over a single USB interface. The motors contain internal controllers which are set to velocity control mode. 

\subsection{Camera}

Next, we employ the wide-angle See3CAM\_24CUG camera which provides a USB3 interface. It is placed 20cm above the center of the labyrinth plate and is mounted to the base of the labyrinth through the use of two aluminum slot profiles such that each frame captures a view of the entire labyrinth plate. The camera outputs a 1920x1200 RGB image at 55 frames per second.




\subsection{Computing}

The motors and camera are connected to a desktop workstation with an AMD Threadripper Pro 5955WX CPU and 2x RTX4090 GPUs. All computational workload, including sensing, actuating, and learning, is performed by said workstation. All software is written using ROS2 \cite{macenski2022robot} to allow for easy intra-process communication and separation of workloads.



%% file: sections/method.tex
\begin{figure*}[ht]
    \centering
    \begin{tikzpicture}

        \pgfdeclarelayer{background}
        \pgfsetlayers{background,main}

        
        \def\height{5.5}
        \def\width{\textwidth}
        \def\labelspace{8pt}
        \coordinate (acting1) at (0, 0);
        \coordinate (acting2) at (0.95*\width, \height);

        \coordinate (learning1) at (0.05*\width, 0.95*\height);
        \coordinate (learning2) at (\width, 1.5*\height);
        \begin{pgfonlayer}{background}
            \draw[rounded corners=8pt, fill=red!7!white, draw=none] (learning1) rectangle (learning2);
            \node at ($(learning1) + (0, 0.55*\height)$) [anchor=north west, inner sep=6pt] {\large\textbf{LEARNING}};
        \end{pgfonlayer}

        \node (replay) at ($(learning1) + (4, 0.35*\height)$) [label={[name=replaylabel]below:Replay buffer}] {{\includegraphics[height=1cm]{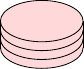}}};
        \begin{pgfonlayer}{background}
            \node (replayfit) [fill=red!30, draw=none, rounded corners=5pt, fit={(replay) (replaylabel)}] {};
        \end{pgfonlayer}

        \node (aug) at ($(replay) + (3.1, 0)$) [label={[name=auglabel]below:Augmentation}] {{\includegraphics[height=1cm]{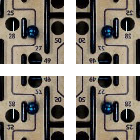}}};
        \begin{pgfonlayer}{background}
            \node (augfit) [fill=red!30, draw=none, rounded corners=5pt, fit={(aug) (auglabel)}] {};
            \node [fill=red!15, fit={(aug)}, rounded corners=5pt, inner sep=-1pt] {};
        \end{pgfonlayer}

        \node (world) at ($(aug) + (3, 0)$) [label={[name=worldlabel, rotate=90, anchor=base]left:{\footnotesize{model}}}] {{\includegraphics[height=1cm]{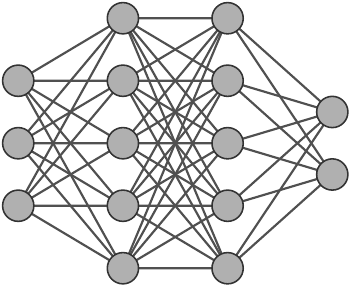}}};
        \node (actor) at ($(world) + (2.5, 0)$) [label={[name=actorlabel, rotate=90, anchor=base]left:{\footnotesize{actor}}}] {{\includegraphics[height=1cm]{img/nn.pdf}}};
        \node (critic) at ($(actor) + (2, 0)$) [label={[name=criticlabel, rotate=90, anchor=base]left:{\footnotesize{critic}}}] {{\includegraphics[height=1cm]{img/nn.pdf}}};

        \node [draw=none, rounded corners=5pt, fit={(world) (critic)}, label={[name=dreamerlabel]below:\phantom{y}Model-Based RL\phantom{y}}, inner sep=0pt, outer sep=0pt] {};
        \begin{pgfonlayer}{background}
            \node (dreamerfit) [fill=red!30, draw=none, rounded corners=5pt, fit={(world) (worldlabel) (dreamerlabel) (critic)}] {};
            \node (modelfit)[fill=red!15, fit={(world) (worldlabel)}, rounded corners=5pt, inner sep=-1pt] {};
            \node (actcriticfit)[fill=red!15, fit={(actorlabel) (critic)}, rounded corners=5pt, inner sep=-1pt] {};
        \end{pgfonlayer}

        \begin{pgfonlayer}{background}
            \draw[rounded corners=8pt, fill=blue!7!white, draw=none] (acting1) rectangle (acting2);
            \node at ($(acting1) + (0.95*\width, 0)$) [anchor=south east, inner sep=6pt] {\large\textbf{ACTING}};
        \end{pgfonlayer}

        \node (real) at ($(acting1)+(0.12*\width, 0.5*\height)$) [fill=blue!30, rounded corners=5pt, align=center, draw=none]{{\includegraphics[width=2.8cm]{img/cyberrunner_setup.png}}\\Physical system};

        \node (img) at ($(real.north)+(4,-0.12)$) [label={[name=imglabel]below:{Camera image}}, anchor=north] {{\includegraphics[width=3cm]{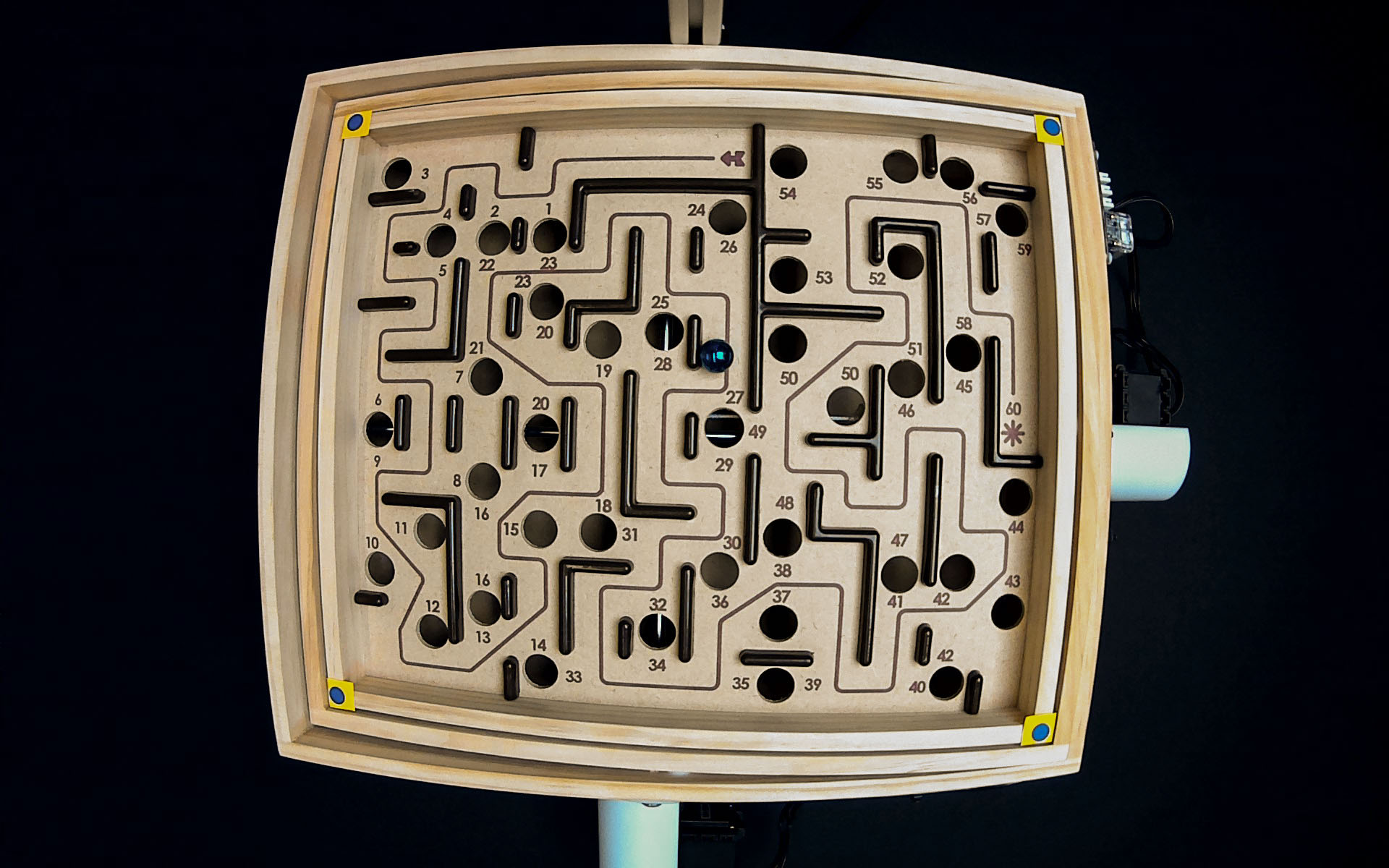}}};
        \begin{pgfonlayer}{background}
            \node (imgfit) [fill=blue!30, fit={(img) (imglabel)}, rounded corners=5pt] {};
        \end{pgfonlayer}

        \node (vec) at ($(img.north)+(5.7, 0)$) [label={[name=veclabel, rotate=90, anchor=base]left:{\footnotesize vec obs}}, anchor=north] {$\begin{bmatrix}
            \text{\footnotesize ball position}\\
            \text{\footnotesize plate angles}\\
            \text{\footnotesize path direction}
        \end{bmatrix}$};
        \node (crop) at ($(vec)+(0, -1.6)$)[label={[name=croplabel, rotate=90, anchor=base]left:{\footnotesize img obs}}] {{\includegraphics[width=1.25cm]{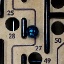}}};
        \node [fit={(crop) (croplabel)}, label={[name=obs]below:Observation}] {};
        \begin{pgfonlayer}{background}
            \node (obsfit) [fill=blue!30, draw=none, rounded corners=5pt, fit={(vec) (veclabel) (crop) (croplabel) (obs)}] {};
            \node[fill=blue!15, fit={(veclabel) (vec)}, rounded corners=5pt, inner sep=-1pt] {};
            \node (cropfull) [fill=blue!15, fit={(croplabel) (crop)}, rounded corners=5pt, inner sep=-1pt] {};
        \end{pgfonlayer}


        \node (rew) at ($(vec.north) + (-2.75, 0)$) [label={[name=rewlabel]below:{Reward}}, anchor=north] {{\includegraphics[width=1.25cm]{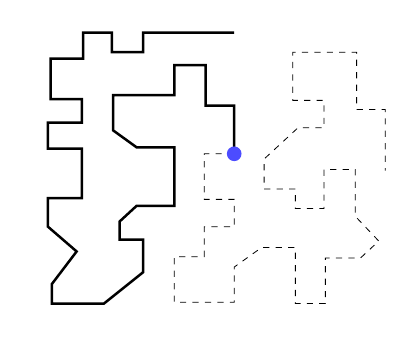}}};
        \begin{pgfonlayer}{background}
            \node (rewfit) [fill=blue!30, draw=none, rounded corners=5pt, fit={(rew) (rewlabel)}] {};
            \node[fill=blue!15, fit={(rew)}, rounded corners=5pt, inner sep=-1pt] {};
        \end{pgfonlayer}

        \node (pol) at ($(vec.north) + (3.2, 0)$) [label={[name=pollabel]below:Recurrent policy}, anchor=north] {{\includegraphics[width=2cm]{img/nn.pdf}}};
        \begin{pgfonlayer}{background}
            \node (polfit) [fill=blue!30, draw=none, rounded corners=5pt, fit={(pol) (pollabel)}] {};
            \node[fill=blue!15, fit={(pol)}, rounded corners=5pt, inner sep=-1pt] {};
        \end{pgfonlayer}

        \draw[-{Latex[round]}, very thick, black!70] (real.east |- imgfit) -- (imgfit);
        \coordinate (obslower) at ($(obsfit.west) - (0, 0.5)$);
        \draw[-{Latex[round]}, very thick, black!70] (imgfit.east |- obslower) -- (obslower);
        \draw[-{Latex[round]}, very thick, black!70] (imgfit.east |- rewfit) -- (rewfit);
        \draw[-{Latex[round]}, very thick, black!70] (obsfit.east |- polfit) -- (polfit);
        \coordinate (reallower) at ($(real.east) - (0, 2)$);
        \draw[-{Latex[round]}, very thick, black!70, rounded corners=5pt] (polfit) |- (reallower);

        \coordinate (mid) at ($0.5*(replayfit.south) + 0.5*(rewfit.north)$);
        \draw[very thick, black!70, rounded corners=5pt] (rewfit.north) -- (rewfit.north |- mid) -- (replayfit.south |- mid) -- (replayfit.south);
        \draw[very thick, black!70, rounded corners=5pt] (obsfit.north) -- (obsfit.north |- mid) -- (replayfit.south |- mid) -- (replayfit.south);
        \draw[-{Latex[round]}, very thick, black!70, rounded corners=5pt] (polfit.north) -- (polfit.north |- mid) -- (replayfit.south |- mid) -- (replayfit.south);

        \draw[-{Latex[round]}, very thick, black!70, rounded corners=5pt] (replayfit.east |- augfit.west) -- (augfit.west);
        \draw[-{Latex[round]}, very thick, black!70, rounded corners=5pt] (augfit.east |- dreamerfit.west) -- (dreamerfit.west);

        \draw[-{Latex[round]}, very thick, black!70, rounded corners=5pt] (dreamerfit.east) to [out=-30,in=30] (polfit.east);

        \draw[-{Latex[round]}, very thick, black!40, rounded corners=5pt] (modelfit) -- (actcriticfit);

    \end{tikzpicture}
    \caption{Overview of the proposed method. On the real-world system, rollouts are collected using the current policy, where observations and reward signals are extracted from the camera images. The collected rollouts are added to the replay buffer using which a model-based RL algorithm aims to optimize a recurrent policy. The learning and acting take place in parallel, and the replay buffer and policy are synchronized between the two processes at the end of each episode.}
    \label{fig:method}
\end{figure*}

\section{Method}
\label{sec:method}
We describe our method in three distinct steps. First, the problem of solving the labyrinth is posed as an infinite-horizon optimal control problem under partial observability using a recurrent policy. Second, we detail how relevant observations and reward information are extracted from the camera images. Lastly, we use model-based RL and data augmentation to optimize the learning objective in a sample-efficient manner. \Cref{fig:method} shows on overview of our method.

\subsection{Problem Statement}
\label{sec:problem}

\newcommand{\idxotk}{
\hspace{-0.5pt}(0\scalebox{0.85}[1.0]{:}k)\hspace{0.5pt}
}

\newcommand{\idxotkmo}{
\hspace{-0.5pt}(0\scalebox{0.85}[1.0]{:}k\scalebox{0.75}[1.0]{\(-\)}1)\hspace{0.5pt}
}

Our aim is to use the presented system to complete the labyrinth as quickly as possible. 
We pose this problem as an infinite-horizon optimal control problem. Formally, the goal is to find a recurrent policy \begin{align}
	\pi \colon (\mathbf{o}\idxotk,\ \mathbf{a}\idxotkmo) \to \mathbf{a}(k)
\end{align}
that is optimal in the sense of maximizing the expected sum of discounted rewards, i.e.,
\begin{align}
	\pi &= \max_{\pi} \mathbb{E}\left[ \sum_{k=0}^{\infty}{\gamma^k r\left({\mathbf{o}}(k), \mathbf{a}(k), \mathbf{o}(k+1)\right) } \right],
\end{align}
where $\gamma$ is the discount factor, $k$ is the discrete time step, $\mathbf{o}(k)$ is the observation at time $k$, $\mathbf{a}(k)$ is the control action to take at time $k$, and $r(k)$ is the scalar reward at time $k$.

As described in \Cref{sec:hardware}, the two employed motors that control the inclination angles of the plate use an internal controller to track angular velocity commands. The control action $\mathbf{a}(k)$ thus consists of the two angular velocity commands for the two motors.

So as to define the reward, we first define a path through the labyrinth by means of defining a sequence of waypoints that are connected linearly. The waypoints are chosen such that the resulting path closely matches the path drawn on the labyrinth board (see \Cref{fig:brio}). 
Next, we define the reward to be the progress along this path at each time step, i.e.,
\begin{align}
    r\left({\mathbf{o}}(k), \mathbf{a}(k), \mathbf{o}(k+1)\right)&= l(k+1)-l(k),
\end{align}
where $l(k)$ is the distance along the path of the point on the path nearest to the ball position at time $k$. The chosen reward encourages the policy to progress through the labyrinth as quickly as possible. Moreover, it is important to note that the chosen reward does not penalize any deviations of the ball from the path as the optimal trajectory of the ball does not necessarily coincide with the given path.

\def\ovec{\mathbf{o}_{\text{vec}}}
\def\oimg{\mathbf{o}_{\text{img}}}

Additionally, we identify two scenarios that the policy should avoid. First, the ball may not fall down any of the holes. Second, the policy may not cheat by skipping over certain parts of the labyrinth. When either of those two scenarios occurs, the corresponding state is set to be a terminal state for which the sum of future rewards is set to 0.

\begin{figure}
    \centering
    \tikzset{>=latex}
    \begin{tikzpicture}

        \coordinate (o) at (0, 0);
        \coordinate (b1) at (-3, -2);
        \coordinate (b2) at (-0.5, -1);
        \coordinate (a1) at (-4, 0);
        \coordinate (a2) at (-3, -0.5);

        \draw[->, dashed, thick] (-3, -2) -- (3, 2);
        \draw[->, dashed, thick] (-4, 0) -- (4, 0);

        \draw (-3, -0.5) -- (3, 0.5);
        \draw (-0.5, -1) -- (0.5, 1);

        \draw (-0.5-3, -1-0.5) -- (0.5-3, 1-0.5);
        \draw (-0.5+3, -1+0.5) -- (0.5+3, 1+0.5);

        \draw (-3-0.5, -0.5-1) -- (3-0.5, 0.5-1);
        \draw (-3+0.5, -0.5+1) -- (3+0.5, 0.5+1);

        \draw[->, very thick, blue] (0, 0) -- (1.4*1, 1.4*0.1666);
        \draw[->, very thick, blue] (0, 0) -- (0.6*0.5, 0.6*1);
        \draw[->, very thick, blue] (0, 0) -- (1.4*-0.1666, 1.4*1);

        \draw[->, very thick, green] (0, 0) -- (1.5, 0);
        \draw[->, very thick, green] (0, 0) -- (0, 1.5);
        \draw[->, very thick, green] (0, 0) -- (0.75, 0.5);

        \draw[->, very thick] (3, -1) -- node [right,midway] {$\displaystyle g$} (3, -2);

        \draw[->, very thick, red] (0, 4) -- (-1.5, 4);
        \draw[->, very thick, red] (0, 4) -- (0, 2.5);
        \draw[->, very thick, red] (0, 4) -- (0.75, 4.5);

        \draw[rounded corners=1pt, fill=black] (-0.2, 4) rectangle (0.2, 4.6);
        \draw (-0.15, 4) -- (-0.5, 3.7) -- (0.5, 3.7) -- (0.15, 4);
        \node at (-1, 4.5) {Camera};

        \draw[thick] (0.75*3+0.4*0.5, 0.75*0.5+0.4*1) -- (0.95*3+0.4*0.5, 0.95*0.5+0.4*1) -- (0.95*3+0.6*0.5, 0.95*0.5+0.6*1) -- (0.75*3+0.6*0.5, 0.75*0.5+0.6*1) -- (0.75*3+0.4*0.5, 0.75*0.5+0.4*1);
        \draw[->] (0.95*3+0.5*0.5, 0.95*0.5+0.5*1) -- (0.95*3+0.5*0.5 + 0.75, 0.95*0.5+0.5*1) -- (0.95*3+0.5*0.5 + 0.75, 3) -- (2.625, 3);
        \node (img) at (2, 3) {{\includegraphics[width=1.25cm]{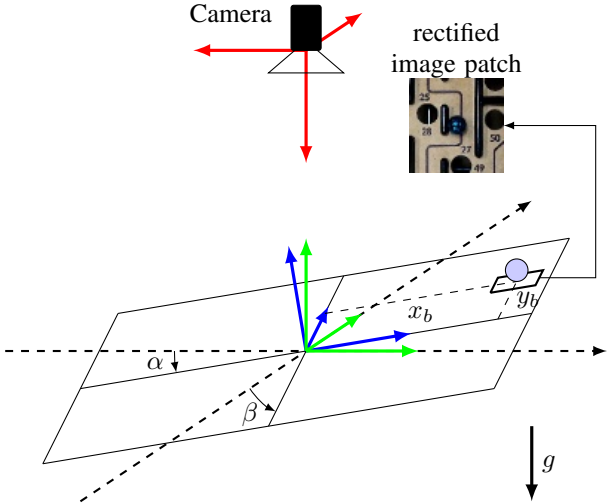}}};
        \node[above of=img, text width=2cm, text centered] {rectified image patch};

        \draw[fill=blue!20] (0.85*3+0.5*0.5, 0.85*0.5+0.5*1+0.15) circle (0.15);
        \draw[dashed] (0.85*3+0.0*0.5, 0.85*0.5+0.0*1) -- node [right,midway] {$\displaystyle y_b$} (0.85*3+0.5*0.5, 0.85*0.5+0.5*1);
        \draw[dashed] (0.0*3+0.5*0.5, 0.0*0.5+0.5*1) -- node [below,midway] {$\displaystyle x_b$}(0.85*3+0.5*0.5, 0.85*0.5+0.5*1);

        \pic [draw, ->, "$\displaystyle\alpha$", angle radius=1.75cm, angle eccentricity=1.15] {angle = a1--o--a2};
        \pic [draw, ->, "$\displaystyle\beta$", angle radius=0.9cm, angle eccentricity=1.25] {angle = b1--o--b2};
        
    \end{tikzpicture}
    \caption{A diagrammatic overview of the labyrinth and the extracted observations.}
    \label{fig:obs}
\end{figure}

From the chosen observations we need to be able to recover a Markovian representation of the state. In the absence of any walls or holes, the labyrinth can be interpreted as a ball-and-plate system \cite{nokhbeh2011modelling}, where the system's state can be defined as
\begin{align}
    \mathbf{x}_{\text{bp}}&=\begin{bmatrix}x_b & \dot{x}_b & y_b & \dot{y}_b & \alpha & \dot{\alpha} & \beta & \dot{\beta}\end{bmatrix}^T,
\end{align} where $(x_b,\ y_b)$ is the ball's position within the plate frame, and $\alpha$ and $\beta$ are the inclination angles of the plate along the two fixed axes of rotation (see \Cref{fig:obs}). 
Since the recurrent policy is conditioned on all past observations and control actions, linear and angular velocities can be recovered from the positional information alone. We thus include only the ball's position and the inclination angles in the observation. Furthermore, we encode the direction of the path through the labyrinth with $n_p=5$ equally spaced points along the path in the direction towards the goal starting at the point on the path closest to the current ball position. While two points are theoretically sufficient to encode the path direction, we find that setting $n_p=5$ yields slight improvements in the sample efficiency of the policy learning.
In summary, the low-dimensional vector observations are given by
\begin{align}
    \ovec&= \begin{bmatrix}
        x_b & y_b & \alpha & \beta & \mathbf{p}_{1:n_p}
    \end{bmatrix}^T,\label{eq:ovec}
\end{align}
where $\mathbf{p}_{1:n_p}$ are the $n_p$ points sampled from the path.

To include information about the position of walls and holes at the current position within the labyrinth, we use a top-down view of a $6\text{cm}\times 6\text{cm}$ window of the labyrinth centered at the ball position. The choice of the window size is an exercise in compromise. Too small a window, and insufficient information is retained, too large a window, and irrelevant information gets encoded into the observation, rendering the task of learning an effective policy more difficult. The final observation is then given by the tuple
\begin{align}
    \mathbf{o}(k)= \left( \ovec(k), \oimg(k)\right), \label{eq:o}
\end{align}
where $\oimg(k)$ denotes the view of the labyrinth surrounding the ball.







\subsection{Image Processing}

The reward and the observations defined above are extracted from the current camera image at each time step. 
In a first step, a camera model is obtained using a state-of-the-art calibration technique \cite{scaramuzza}. The calibration yields the projection function of 3D points to 2D pixel coordinates (and the inverse), as well as the position and orientation of the camera center w.r.t. the labyrinth base.

To facilitate the estimation process, we (i) use a blue-colored steel ball (same dimension, weight, and material as the original ball), and (ii) place blue markers on the corners of the labyrinth plate. Blue is chosen as it is in great contrast to the mostly green and red dominant color of the labyrinth.

The extraction of the observation $\mathbf{o}(k)$ from a given camera image can then be summarized in the following steps:
\begin{enumerate}
    \item The ball and the markers placed on the corners of the plate are tracked using predictive windows centered on the previous estimate of their positions. Within those windows, the ball and markers are detected by masking the image using hand-tuned hue, saturation, value (HSV) ranges, extracting contours from the masked image, and finally computing the center of mass of the extracted contours.
    \item Using the pixel coordinates of the tracked corner markers, together with the calibrated camera model and the knowledge of the position of the corners w.r.t. the labyrinth plate, we can estimate the inclination angles $\alpha$ and $\beta$ of the labyrinth plate using infinitesimal plane-based pose estimation \cite{collins2014infinitesimal}.
    \item Using the pixel coordinates of the tracked ball, together with the pose of the labyrinth plate and the calibrated camera model, an estimate of the ball position $(x_b,\ y_b)$ in the labyrinth plate frame is obtained.
    \item Using the ball position, the pose of the labyrinth plate, and the calibrated camera model, the rectified view of the labyrinth layout surrounding the ball (as defined in \Cref{sec:problem}) is extracted. The view is encoded as a $64\times64$ RGB image.
    \item From the estimated ball position, the point on the pre-defined path closest to the current ball position is obtained, yielding the current progress $l(k)$ along the labyrinth path.
\end{enumerate}


\subsection{Learning and Data Augmentation}

In order to learn the policy, we employ the model-based RL algorithm DreamerV3 \cite{hafner2023dreamerv3} with default hyperparameters. There, a world model learns a Markovian representation of the system as well as the system dynamics. The learned world model is used to generate imagined trajectories which are used to train actor and critic networks. Akin to \cite{wu2023daydreamer}, we deploy DreamerV3 in an asynchronous manner; the learning of the world model, the critic, and the actor, is performed in parallel to acting and collecting experience on the real system. At the end of each episode on the physical setup, policy networks are synced, and the collected experience from the episode is added to the replay buffer.


\begin{figure}
    \centering
    \begin{tikzpicture}
        \node[align=center] at (0.125\columnwidth, 0) {\includegraphics[width=0.22\columnwidth]{img/subimg.jpg}\\\small{Original}};
        \node[align=center] at (0.375\columnwidth, 0) {\includegraphics[width=0.22\columnwidth]{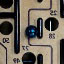}\\\small{Horizontal flip}};   
        \node[align=center] at (0.625\columnwidth, 0) {\includegraphics[width=0.22\columnwidth]{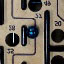}\\\small{Vertical flip}};
        \node[align=center] at (0.875\columnwidth, 0) {\includegraphics[width=0.22\columnwidth]{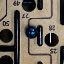}\\\small{180 rotation}};
    \end{tikzpicture}
    \caption{Observations are augmented during training by mirroring about two planes of symmetry.}
    \label{fig:aug}
\end{figure}

To improve the sample efficiency of the learning process, sequences sampled from the replay buffer during training are augmented by randomly mirroring the trajectories about the two vertical planes that coincide with the two axes of rotation. This includes mirroring both the low-dimensional and the image observations, as well as the control actions. In essence, we obtain trajectories on vertically and/or horizontally flipped views of the labyrinth (see \Cref{fig:aug}), yielding four distinct versions of the labyrinth. We argue that learning from mirrored versions of the labyrinth translates to a more robust policy that is able to generalize to unseen parts of the labyrinth.


%% file: sections/results.tex
\section{Results}
\label{sec:results}

The validity of the methods presented is verified by (i) running a single training run on the physical system and evaluating the performance of the policy, and (ii) performing an ablation study using a simulated version of the labyrinth. 


    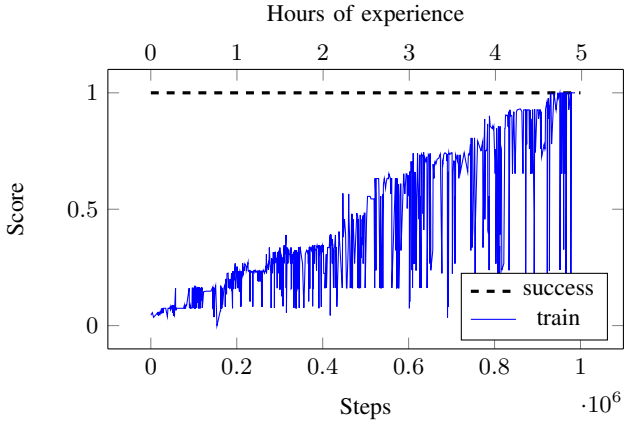
\begin{figure}
        \centering
        \begin{tikzpicture}
        \tikzstyle{every node}=[font=\small]
        \def\maxv{2.462}
        \begin{axis}[
            y filter/.code={\pgfmathparse{#1/\maxv}\pgfmathresult},
            no marks,
            axis x line*=bottom,
            height=5.3cm,
            width=\axisdefaultwidth,
            xlabel={Steps},
            ylabel={Score},
            legend pos=south east
        ]
        \addplot [domain=0:1e6, samples=100, dashed, style=very thick, black]{\maxv};
        \addplot [blue] table [x=Step, y=Value, col sep=comma] {img/scores.csv};
        \legend{success, train};
        \end{axis}
        \begin{axis}[
          x filter/.code={\pgfmathparse{#1/55.0/3600}\pgfmathresult},
          axis x line*=top,
          height=5.3cm,
          width=\axisdefaultwidth,
          xlabel={Hours of experience},
          xlabel near ticks,
          axis y line=none,
          ]
          \addplot [draw=none] table [x=Step, y=Value, col sep=comma] {img/scores.csv};
        \end{axis}
        \end{tikzpicture}        
        \caption{Single training run on the physical system. A policy that successfully completes the labyrinth is learned within 5 hours. The $x$-axis represents the number of environment interactions, while the $y$-axis represents the normalized accumulated reward for each episode.}
        \label{fig:curve-real}
    \end{figure}



The learning of the policy on the real-world labyrinth is conducted with a budget of $1$ million time steps (equivalent to $5.05\si{\hour}$ at a control rate of $55\si{\hertz}$). During learning on the physical system, we manually place the ball back to its starting position by hand whenever an episode terminates. The normalized accumulated reward for each episode during training is plotted in \Cref{fig:curve-real}. As can be seen, the policy successfully navigates the ball to the end of the labyrinth with less than $5$ hours of collected data.

\begin{figure}
    \centering
    \includegraphics[width=0.9\columnwidth]{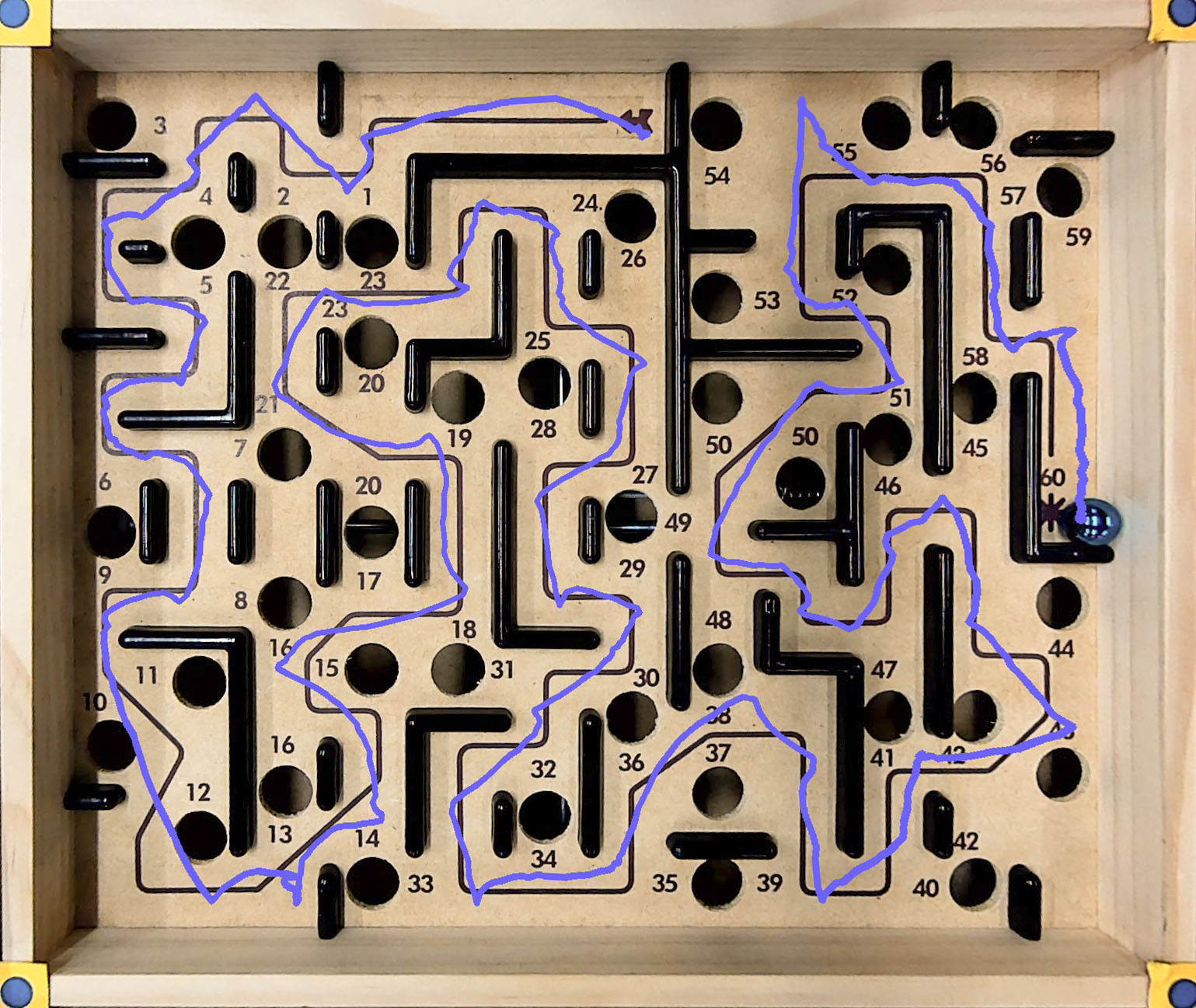}
    \caption{A rectified view of the ball trajectory through the labyrinth during a successful run of the final policy.}
    \label{fig:realrun}
\end{figure}

\begin{table}
    \centering
    \renewcommand{\arraystretch}{1.3}
    \caption{Comparison between our method and the previous fastest recorded completion time.}
    \begin{tabular}{lcc}
        \hline
         Method & Success rate &  Completion time\\\hline
         This paper ($N=50$) &  $76\%$ & $15.73\si{\second}\pm 0.36\si{\second}$ \\
         Human (best)\footnotemark & - & $15.95\si{\second}$\\\hline
    \end{tabular}
    \label{tab:res}
\end{table}

Additionally, we run the resulting policy on the physical setup $N=50$ times and report the success rate and the average completion time for successful runs in \Cref{tab:res}. On average, the policy outperforms the previously fastest recorded time by $0.22\si\second$ and does so with a $76\%$ success rate. \Cref{fig:realrun} illustrates a successful run of the final policy. As can be seen, the policy effectively exploits the walls to quickly redirect the ball. It can also be observed that the final policy navigates the ball in the close vicinity of the holes in order to maximize its performance (such as near holes 10 and 12 in \Cref{fig:realrun}). We found that this greediness of the policy leads to failed runs and that training the policy for more than the allocated $1$ million time steps increases the robustness and success rate of the policy.

\footnotetext[1]{According to \url{https://recordsetter.com/world-record/complete-game-labyrinth/12617}.}

    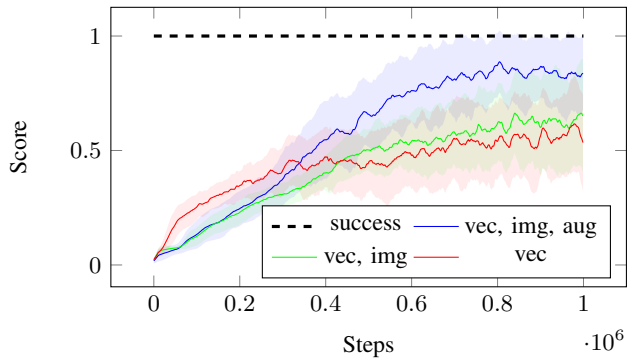
\begin{figure}
        \centering
        \begin{tikzpicture}
        \tikzstyle{every node}=[font=\small]
        \def\maxv{2.462}
        \begin{axis}[
            y filter/.code={\pgfmathparse{#1/\maxv}\pgfmathresult},
            no marks,
            height=5.3cm,
            width=\axisdefaultwidth,
            xlabel={Steps},
            ylabel={Score},
            legend pos=south east,
            legend columns=2
        ]
        \addplot [domain=0:1e6, samples=100, dashed, style=very thick, black]{\maxv};
        \addplot [name path=brio-upper,draw=none] table[x=step, y expr=\thisrow{brio}+\thisrow{brio_err}, col sep=comma] {img/brio_sim_curves.csv};
        \addplot [name path=brio-lower,draw=none] table[x=step, y expr=\thisrow{brio}-\thisrow{brio_err}, col sep=comma] {img/brio_sim_curves.csv};
        \addplot [fill=blue, opacity=0.08] fill between[of=brio-upper and brio-lower];
        \addplot [name path=uniform-upper,draw=none] table[x=step, y expr=\thisrow{uniform}+\thisrow{uniform_err}, col sep=comma] {img/brio_sim_curves.csv};
        \addplot [name path=uniform-lower,draw=none] table[x=step, y expr=\thisrow{uniform}-\thisrow{uniform_err}, col sep=comma] {img/brio_sim_curves.csv};
        \addplot [fill=green, opacity=0.08] fill between[of=uniform-upper and uniform-lower];
        \addplot [name path=state-upper,draw=none] table[x=step, y expr=\thisrow{state}+\thisrow{state_err}, col sep=comma] {img/brio_sim_curves.csv};
        \addplot [name path=state-lower,draw=none] table[x=step, y expr=\thisrow{state}-\thisrow{state_err}, col sep=comma] {img/brio_sim_curves.csv};
        \addplot [fill=red, opacity=0.08] fill between[of=state-upper and state-lower];
        \addplot [blue] table [x=step, y=brio, col sep=comma] {img/brio_sim_curves.csv};
        \addplot [green] table [x=step, y=uniform, col sep=comma] {img/brio_sim_curves.csv};
        \addplot [red] table [x=step, y=state, col sep=comma] {img/brio_sim_curves.csv};
        \legend{success,,,,,,,,,, {vec, img, aug}, {vec, img}, vec};
        \end{axis}
        \end{tikzpicture}
        \caption{The mean and standard deviation of the scores averaged over 16 seeds for variations of our method incorporating different components. "vec" denotes the low-dimensional observation $\ovec$, "img" denotes the use of the image patch $\oimg$, and "aug" denotes the usage of the proposed data augmentation. The addition of the image patch and data augmentation leads to a significant increase in the performance of the policy. The $x$-axis represents the number of environment interactions, while the $y$-axis represents the normalized accumulated reward for each episode.}
        \label{fig:curve-sim}
    \end{figure}

To prove the merits of different aspects of the proposed method, we run ablation studies using a simulated version of the labyrinth; we recreate the system, including the labyrinth, the motors, and the camera, in simulation using the MuJoCo \cite{todorov2012mujoco} physics engine. Then, three different training settings are assessed in simulation; (i) using only the vector observations as described in \Cref{eq:ovec} without data augmentation, (ii) using both vector and image observations (\Cref{eq:o}) without data augmentation, and (ii) the full proposed method including vector and image observations as well as random mirroring of the observations during training. For each scenario, the learning is performed across 16 different seeds, with the same hyperparameters as used on the physical system. The results are shown in \Cref{fig:curve-sim}. As can be seen, the addition of each component leads to a higher final performance given the same number of environment interactions. As mentioned in \Cref{sec:method}, we interpret this result as follows. The image patch observations encode the position of the walls and the holes near the ball, helping the policy make more informed decisions, while the data augmentation significantly improves sample efficiency by diversifying the training data and improving generalization to previously unseen parts of the labyrinth. While the above-mentioned ablation study could in theory be performed on the physical system, the time investment required to run the three scenarios multiple times in the real world is deemed too costly.

%% file: sections/conclusion.tex
\section{Conclusion}
\label{sec:conc}

In this work, we have presented a novel approach to solving a popular and widely-available real-world labyrinth game, leveraging state-of-the-art deep reinforcement learning techniques. Our method demonstrates remarkable proficiency in navigating the labyrinth, achieving completion times faster than any previously recorded. Furthermore, the combination of appropriately chosen observations that encode all relevant information, model-based RL, and data augmentation exploiting inherent symmetries of the system, result in a sample-efficient learning method that is directly employed on the real system.




Looking further ahead, we aim to open-source the hardware and software of the robotic system so as to provide other researchers the opportunity to utilize the labyrinth as a testbed, as we believe it provides a great real-world benchmark with low space requirements, low hardware complexity, and modest cost, for both model-based control as well as reinforcement learning.


